\documentclass[10pt,a4paper,twocolumn]{article}
\title{Experiments with adversarial attacks on text genres}
\author{Mikhail Lepekhin, Serge Sharoff\\MIPT, Moscow; University of Leeds, UK}
\date{}
\usepackage[british]{babel}
\usepackage{verbatim}
\usepackage{graphicx}
\usepackage{caption}
\usepackage{todonotes}
\usepackage{enumitem}
\usepackage{times}
\usepackage{hyperref}
\usepackage{apalike}
\usepackage{pdflscape}
\usepackage{amsmath}
\usepackage[T1]{fontenc}

\begin{document}
\maketitle

\begin{abstract}
Neural models based on pre-trained transformers, such as BERT or XLM-RoBERTa, demonstrate SOTA results in many NLP tasks, including non-topical classification, such as  genre identification.  However, often these approaches exhibit low reliability to minor alterations of the test texts. A related probelm concerns topical biases in the training corpus, for example, the prevalence of words on a specific topic in a specific genre can trick the genre classifier to recognise any text on this topic in this genre. In order to mitigate the reliability problem, this paper investigates techniques for attacking genre classifiers to understand the limitations of the transformer models and to improve their performance. While simple text attacks, such as those based on word replacement using keywords extracted by tf-idf, are not capable of deceiving powerful models like XLM-RoBERTa, we show that embedding-based algorithms which can replace some of the most ``significant'' words with words similar to them, for example, TextFooler, have the ability to influence model predictions in a significant proportion of cases. 
\end{abstract}

\section{Introduction}

Non-topical text classification concerns a wide range of problems that are aimed at predicting a text property that is not connected directly to the text topic, for example, at predicting its genre, difficulty level, the age or the first language of its author, etc.  Unlike topical text classification, non-topical text classification needs a model that predicts a label on the basis of its stylistic properties.  Automatic genre identification is one of the standard problems of non-topical text classification, as it is useful in many areas such as information retrieval, language teaching or basic linguistic research \cite{santini10genreintro}.

An early comparison of various datasets, models and linguistic features for genre classification \cite{sharoff10lrec} shows that traditional machine learning models, for example, SVM, can be very accurate in genre classification on their native dataset, but suffer from a dataset shift. Since then, many new approaches to text classification have emerged.  In particular, BERT (Bidirectional Encoder Representations from Transformers) is an efficient pre-trained model based on the Transformer architecture \cite{bert}.  It achieves the state-of-the-art results for various NLP tasks, including text classification. In this study we use XLM-RoBERTa \cite{xlm_roberta} is an improved variant of BERT. It has the same architecture, but uses bigger and more genre diverse corpora and an updated pre-training procedure. In addition, XLM-RoBERTa is a multilingual model trained on Common Crawl data in comparison to multilingual BERT only trained on Wikipedia.

One of the most significant problems in genre classification concerns topical shifts \cite{petrenz10}.  If in the training corpus a specific topic is more frequent for a specific genre, then many classification models can be biased towards indicating this genre by the keywords of this topic.  This becomes especially problematic in the case of data shift between the training and testing corpora \cite{petrenz10}.  For this reason, they check reliability of their genre classifiers by testing on datasets from different domains.  We test this in our study too.

\begin{table*}[t!]
  \centering
  \setlength{\tabcolsep}{3pt}
\begin{tabular}{llrrrr||rl|r}
Genre label & Prototypes & \multicolumn{2}{c}{FTD EN} & \multicolumn{2}{c}{FTD RU} &
\multicolumn{2}{c}{Natural annotation EN} &
\multicolumn{1}{c}{LJ RU} \\
            &    & Train &  Val  & Train & Val & Test & Sources & Test  \\
\hline
Argument 	& Expressing opinions, editorials 	& 276 	& 77 & 207 & 77  & 400 & \cite{kiesel19semeval} & 481 \\
Fiction 	& Novels, songs, film plots 	& 69	& 28 & 62 & 23 & 400 & BNC\&Brown & 199\\
Instruction 	& Tutorials, FAQs, manuals 	& 141	& 50 & 59 & 17 & 400 & StackExchange & 384\\
News    	& Reporting newswires 	& 114 	& 37 & 379 & 103 & 400 & Giga News & 1518\\
Legal   	& Laws, contracts, T\&C 	& 56	& 17 & 69 & 13 & 400 & UK and US legal codes & 14 \\
Personal 	& Diary entries, travel blogs 	& 72 	& 19 & 126 & 49 & 400 & ICWSM & 513\\
Promotion 	& Adverts, promotional postings 	& 218  	& 66 & 222 & 85  & 400 & promo sites & 68\\
Academic 	& Academic research papers 	& 59 	& 23 & 144 & 49 & 400  & arxiv.org & 20 \\
Information 	& Encyclopedic articles 	& 131 & 38 & 72 & 33 & 400 & Wikipedia & 171\\
Review  	& Product reviews 	& 48	& 22 & 107 & 34 & 400 & Amazon reviews & 185\\
\hline
  & \textbf{Total} & 1184 & 377 & 1447 & 483 & 4000 & & 3553 \\
\end{tabular}
\caption{Training and testing corpora \label{tabTraining}}
\end{table*}

There have been numerous attempts to attack NLP models by making minor changes to a text which lead to different predictions.  An overview of different methods is presented in \cite{adversarial_attacks_survey}.  These techniques help to reveal the flaws of the NLP models and to find out what are the features in the texts that are taken into account by the models. 
TextFooler \cite{textfooler} sorts the words of texts under attack by their impact on the target class probability and tries to replace the most important words with their closest neighbours with the similarity defined as the dot product between the corresponding word embeddings. BertAttack \cite{bertattack} has a similar algorithm, but instead of using word embeddings it relies on Bert token embeddings. Because of this, BertAttack processes the whole words and subword tokens in different ways, while trying to find suitable words to replace subword tokens.

Until now, there have been no reports of successful attempts of attacking genre classifiers or non-topical classification in general using neural methods, even though it is important to understand their reliability and to find ways for improving their robustness.  In this study, we test two methods to attack text genre classifiers.  The first method is based on swapping the keywords which are found with tf-df extraction, while the second method applies a modified TextFooler algorithm. Moreover, we try to improve the performance of the original classifiers by adding a set of texts broken by TextFooler to the training corpus.

In this paper we perform the following steps to investigate attacking techniques and to improve the reliability of the genre classifier:

\begin{enumerate}[noitemsep]
\item training a baseline classifier using XLM-RoBERTa (Section~\ref{secBase});
\item attacking the XLM-RoBERTa classifier by swapping topical keywords between the genres (Section~\ref{secKeywords});
\item attacking the XLM-RoBERTa classifier with TextFooler (Section~\ref{secTextFooler});
\item performing targeted attacks on the XLM-RoBERTa classifier  (Section~\ref{secTargetedTextFooler});
\item training a new XLM-RoBERTa classifier by using the original training corpus combined with the successfully attacked texts (Section~\ref{secReliableModel});
\end{enumerate}

All code, data, and materials to fully reproduce the experiments are openly available.\footnote{\url{https://github.com/MikeLepekhin/TextGenresAttack}}

\section{Baseline}
\label{secBase}
  
\begin{table*}[t]
  \centering \small
 \begin{tabular}{cp{0.8\linewidth}} 
 \hline
 \textbf{Genre} & \textbf{Keywords}\\
 \hline
 Argument & united, nations, reconciliation, international, development, people, security, countries  \\
  Fiction & said, would, one, could, little, man, came, like, went, upon \\ 
  Instruction & tap, device, screen, email, tab, select, settings, menu, contact, message \\
  News & said, million, committee, disarmament, kongo, report, program, also, budget, democratic \\
  Legal & shall, article, may, paragraph, court, person, order, department, party, state \\
  Personal & church, one, like, people, could, really, congo, time, years, would \\
  Promotion & viagra, cialis, online, writing, posted, service, levitra, business, buy, essay \\
  Academic & system, quantum, fault, data, software, image, node, faults, application, fig \\
  Information & committee, convention, parties, secretariat, iran, meeting, shall, mines, states, conference \\
  Review & google, home, new, like, star, paul, one, shoes, pro, art \\ 
 \hline
\end{tabular}
 \caption{Examples of English keywords extracted with tf-idf}
 \label{tabKeywords}
\end{table*}

\begin{table}[t]
  \begin{center}
\begin{tabular}{cccc} 
\hline
\textbf{Replaced} & \textbf{10\%} & \textbf{50\%} & \textbf{100\%} \\
\hline
EN & 14 (1.1\%) & 31 (2.5\%) & 196 (15.5\%) \\
RU & 22 (1.5\%) & 44 (3.0\%) & 148 (10.0\%) 
\end{tabular}
  \end{center}
\caption{Successful attacks with keyword replacement}
\label{tabTfIdfAttack}
\end{table}

\subsection{Training data}

For training the genre classifiers, we use existing FTD datasets in English and in Russian \cite{sharoff18genres}. Each of them contains more than 1,500 texts from a wide range of sources annotated with 10 genre labels, see Table~\ref{tabTraining}. The dataset is relatively balanced with the most common categories being Argumentation and Promotion.  For validation of the success of attacking models at the last stage (see the next section) we reserve a small portion of this dataset obtained by stratified sampling (columns Val in~\autoref{tabTraining}), which is not used in the training and attacking pipelines.

It is known that genre classifiers are often not robust when applied to a different corpus with the same labels \cite{sharoff10lrec}, therefore we use independently produced test sets to simulate out-of-domain performance on large collections coming from a smaller number of sources.  This makes them different from the training datasets, which came from a much wider range of sources.

For the Russian test set we use posts from LiveJournal, a social media platform popular in Russia. Each of these texts has been annotated by two assessors. Those texts for which the annotators did not agree have been adjudicated by an expert annotator. Since LiveJournal is a social media platform, the distribution of its texts significantly differs from the FTD  corpora. It contains fewer Legal, Academic and Promotion texts and more News, Personal and Instruction texts (see~\autoref{tabTraining}).

As we lack an independent test set for English, we use ``natural annotation'' in the sense of collecting examples of texts for each genre from sources relatively homogenous with respect to this genres, such as StackExchange, which mainly contains instructive texts, or Wikipedia, which mainly contains texts for reference information, see more details in the Sources column in~\autoref{tabTraining}.  Similar to social media data, natural annotation creates its biases, for example, submissions to arxiv.org tend to be on topics in physics or computer science, so that we will be able to test predictions in the presence of biases. 

\subsection{Training genre classifiers}

We fine-tune the baseline XLM-RoBERTa classifiers following the same architecture as \cite{sun19} using the training part of the FTD corpus for 10 epochs with the Adam optimiser with $\text{learning\ rate}=5\cdot 10^{-5}$ since these hyperparameters are used for fine-tuning in the original papers for several BERT-like models \cite{bert,roberta}.

\section{Genre attacks}
The genre attack task is to make minimal alterations to a target text with the aim to change its prediction by an existing classifier.  If a test text can be altered to change the label predicted by the classifier, and if this can be achieved within a fixed limit of alterations, the text is counted as ``broken''. We can try untargeted and targeted attacks:
\begin{description}
\item [untargeted] these are attacks that intend to force the classifier to change its correct prediction on a test set text to produce any incorrect label from our set of labels;
\item [targeted] the opposite attack direction when we attack texts for which the classifier makes a mistake by making alterations to force the classifier to predict the correct label. 
\end{description}

The genre attacks are conducted to achieve cross-validation for attacks without leaking information about the target texts to the classifier: we randomly shuffle the training dataset and make 5 iterations of the cross-validation mechanism: For every $i$ the texts with numbers from $0.2i|X|$ to $0.2(i+1)|X|-1$ are used as test texts to attack the classifier which has been trained on the remaining texts from the training corpus.  Thus, we get five architecturally identical classifier models with slightly different weights, as well as a set of successfully attacked texts we can use our analysis below.

\begin{table*}[t]
\small
\setlength{\tabcolsep}{4pt}
\begin{tabular}{|p{0.5\linewidth}|p{0.5\linewidth}|} 
\hline
\textbf{Original} & \textbf{Attacked} \\
\hline
As a Company Limited by Guarantee \textbf{this} charity is owned not by any shareholders but by its members. Only members can vote at Annual General Meetings to elect officers and Directors or become Directors of the charity. So if you would like to help us in this way, contributing at least £5 per year and in return receive regular updates and an invitation to the AGM please complete a membership form Company Membership Form Friends Membership Form There is also the option to make a monthly donation towards our work. As little as £2 a month can make a real difference to Emmaus Projects. &
As a Company Limited by Guarantee \textbf{that} charity is owned not by any shareholders but by its members. Only members can vote at Annual General Meetings to elect officers and Directors or become Directors of the charity. So if you would like to help us in this way, contributing at least £5 per year and in return receive regular updates and an invitation to the AGM please complete a membership form Company Membership Form Friends Membership Form There is also the option to make a monthly donation towards our work. As little as £2 a month can make a real difference to Emmaus Projects. \\
\hline
\textbf{label: Promotion} & \textbf{label: Argument}\\
\hline
\end{tabular}
\caption{Untargeted attack example with TextFooler}
\label{tabUntargetedAttack}
\end{table*}

\subsection{Untargeted attack by swapping topical keywords}
\label{secKeywords}

First, we test a simple text attack generator which is based on replacing keywords extracted for each genre by keywords extracted for other genres.  The keywords are defined by their tf-idf scores within the genre texts. \autoref{tabKeywords} lists the most significant keywords according to the  tf-idf score.  Some keywords correspond to their genres quite reasonably, for example, those from Fiction or Legal texts. However, most genres have fairly topical keywords, which indicates the prevalence of specific topics in the training corpus. For example, the keyword lists show that both Argument and News contain a lot of texts about international politics, while many Instruction texts refer to Internet services or communication devices.

Then the attack generator replaces a certain percentage of the keywords for a genre to a keyword of a random genre. We choose the following range of the keywords to be replaced: 10\%, 50\%, 100\%.  Contrary to our expectations concerning the prevalence of topic-specific keywords, our XLM-R classifier is reasonably robust to attacks on both English and Russian texts, as the rate of successfully broken texts is fairly low even when all tf-idf keywords are replaced, see \autoref{tabTfIdfAttack}.

\begin{table*}[t]
  \begin{center}
\begin{tabular}{c|c|c|c|c} 
\hline
USE & \textbf{Language} & \textbf{k=15} & \textbf{k=30} & \textbf{k=50} \\
\hline
0.84 & EN & 416 (32.9\%) & 438 (34.7\%) & 453 (35.8\%) \\
0.84 & RU & 686 (47.4\%) & 718 (49.6\%) & 744 (51.4\%) \\ 
\hline
0.6 & EN & 424 (33.5\%) & 444 (35.1\%) & 457 (36.2\%) \\
0.6 & RU & 687 (47.5\%) & 720 (49.8\%) & 744 (51.4\%) \\ 
\hline
0 & EN & 424 (33.5\%) & 444 (35.1\%) & 457 (36.2\%) \\
0 & RU & 687 (47.5\%) & 720 (49.8\%) & 744 (51.4\%) \\ 
\hline
\end{tabular}
  \end{center}
\caption{Successful untargeted attacks with different USE thresholds}
\label{tabZeroThreshold}
\end{table*}


\begin{table*}[t]
\centering
\setlength{\tabcolsep}{3pt}
{\small
\begin{tabular}{|c|p{0.9\linewidth}|} 
\hline
\textbf{Genre} & \textbf{Words} \\
\hline
Argument & people$\to$residents~(14), have$\to$be~(13), have$\to$has~(12), world$\to$worldwide~(8), be$\to$have~(8), social$\to$societal~(8), do$\to$know~(7), children$\to$infants~(7), people$\to$individuals~(7), nuclear$\to$fissile~(7) \\ 
Fiction & had$\to$has~(12), had$\to$have~(10), will$\to$wants~(10), have$\to$has~(6), king$\to$monarch~(5), each$\to$every~(4), did$\to$does~(4), came$\to$coming~(4), come$\to$happen~(4), have$\to$be~(4) \\
Instruction & do$\to$know~(18), will$\to$wants~(12), be$\to$have~(10), have$\to$be~(10), should$\to$ought~(10), click$\to$clicking~(6), choose$\to$choices~(5), based$\to$inspired~(4), try$\to$trying~(4), example$\to$examples~(4) \\
News & will$\to$want~(13), has$\to$maintains~(7), has$\to$have~(6), be$\to$have~(5), will$\to$wants~(5), have$\to$be~(5), said$\to$stating~(5), year$\to$olds~(4), new$\to$ny~(4), week$\to$days~(4) \\
Legal & be$\to$have~(22), shall$\to$hereof~(18), shall$\to$howsoever~(11), terms$\to$terminology~(8), order$\to$ordering~(8), person$\to$somebody~(8), conditions$\to$situations~(5), contract$\to$agreement~(5), agreement$\to$agreed~(5) \\
Personal & life$\to$lives~(6), do$\to$know~(5), think$\to$suppose~(5), wanted$\to$want~(5), felt$\to$knew~(4), people$\to$individuals~(3), started$\to$begin~(3), went$\to$going~(3), design$\to$styling~(2), so$\to$because~(2) \\
Promotion & be$\to$have~(6), new$\to$ny~(5), business$\to$commerce~(5), company$\to$corporation~(5), have$\to$be~(5), products$\to$byproducts~(4), opportunity$\to$opportunities~(4), help$\to$aid~(4), company$\to$venture~(4), model$\to$models~(4)\\
Academic & scattering$\to$scatter~(8), have$\to$be~(5), findings$\to$confirmatory~(3), mathematical$\to$dynamical~(3), analysis$\to$analyzed~(3), show$\to$showcase~(3), idea$\to$thought~(3), computation$\to$computing~(3), be$\to$have~(3) \\
Information & system$\to$integrator~(4), number$\to$numbering~(4), has$\to$have~(3), system$\to$mechanism~(3), each$\to$every~(3), had$\to$has~(2), person$\to$someone~(2), little$\to$scant~(2), astronomy$\to$ephemeris~(2), ehc$\to$liga~(2) \\
Review & google$\to$yahoo~(3), quality$\to$dependability~(2), review$\to$reassessment~(2), synth$\to$synths~(2), movie$\to$movies~(1), company$\to$corporation~(1), rescue$\to$rescued~(1), get$\to$got~(1), engadget$\to$wired~(1)\\
\hline
\end{tabular}
}
\caption{Most common English word pairs amended with untargeted TextFooler attack}
\label{tabTextFoolerWordPairs}
\end{table*}

\begin{table*}[t]
  \small
\begin{tabular}{|p{0.5\linewidth}|p{0.5\linewidth}|} 
\hline
\textbf{Original} & \textbf{Attacked} \\
\hline
In addition to the internet connection, you \textbf{should} also \textbf{try} to have at least 100 MB of free space available on your drive when you install Titan Poker.  &
In addition to the internet connection, you \textbf{need} also \textbf{trying} to have at least 100 MB of free space available on your drive when you install Titan Poker. \\
\hline
\textbf{label: Instruction} & \textbf{label: Review}\\
\hline
\end{tabular}
\caption{Example of deterioration of grammar in untargeted attack}
\label{tabUngrammarUntargetedAttack}
\end{table*}

\subsection{Attacking with untargeted TextFooler}
\label{secTextFooler}

\begin{table}[t]
\centering
\begin{tabular}{|c|r|r|} 
\hline
\textbf{Genre} & \textbf{English} & \textbf{Russian} \\
\hline
Argument 	&  12.0 & 12.0 \\
Fiction 	&  19.0 & 12.0 \\
Instruction 	&  16.5 & 23.5  \\
News reports 	&  10.0 & 21.0  \\
Legal   	&  26.0 & 20.0 \\
Personal 	&  18.0 &   6.0\\
Promotion 	&  12.0 &  25.0\\
Academic 	&  11.0 &  18.0\\
Information 	&   5.5 &   5.5 \\
Review 		&   3.0 &   3.5\\
\hline
\end{tabular}
\caption{The median number of words per text for successful genre attacks}
\label{tabAttackAccuracy}
\end{table}

The original TextFooler algorithm has the following stages. First, we order the words $w_i$ of the training corpus (after excluding the stop words) by the descending order of their importance scores $I_w$, that defined in the following way:
{\small
\[
    I_{w_i}=
    \begin{cases}
      F_Y(X)-F_Y(X_{\setminus w_i}),\qquad \text{if}\ F(X)=F(X_{\setminus w_i})=Y \\
      (F_Y(X)-F_Y(X_{\setminus w_i}))+(F_{\bar Y}(X_{\setminus w_i})-F_{\bar Y}(X)),\\\qquad\text{if}\ F(X)=Y, F(X_{\setminus w_i})=\bar Y,\text{ and } Y\neq \bar Y
      .
    \end{cases}
  \]
}
where $F(X)$ is the predicted label for text $X$, $F_{Y}(X)$ is the predicted probability of the genre $Y$ for the text $X$, and $X_{\setminus w_i}$ denotes a text with $w_i$ removed.  The intuition of the importance score is that removal of a more important word leads to greater distortion of the predicted probability.

Then for every word in the attacked text, $k$ closest words are chosen by the value of the dot product of their embeddings with the embedding of the original word. These words are the candidates for replacing the original word. We iterate through the words $w_i$ in the order of their importance and try to replace each of them with one of the candidates following rules in a set of filters. If we succeed in doing that, then the text replacement is considered as successful. Otherwise, we continue to iterate through the list of candidates. If we cannot find a candidate $\title{w}_i$ for replacing the word $w_i$, we try the word for which the classifier gives the minimal probability of the original class for the text with this replacement. If we have iterated all over the words $w_i$, but the classifier still predicts the original label for the text, the attack is unsuccessful.

The filters for choosing a suitable replacement can vary. First, we can keep the same part-of-speech tag (usually on the top level of tags, for example, NOUN$\to$NOUN). Second, we can vary the lower limit threshold for the word similarity score for each candidate.  In the original TextFooler algorithm, this threshold is fixed at 0.5. In our study the 20-80 percentile range for the embedding similarities between each word and its \textbf{closest} neighbour is 0.61--0.82 for English and 0.67--0.82 for Russian. 
If we take into account the top-15 most similar embeddings for each word embedding, the 20--80 percentile range for English is 0.49--0.66, for Russian it is 0.52--0.68.  This limits the range of values for selecting the similarity threshold.

Finally, to preserve the meaning and the grammatical correctness of the attacked texts, we estimate the similarity between the original sentence and its attacked version with the Universal Sentence Encoder \cite{use}.  The original TextFooler paper fixed the threshold of the minimal score to 0.84, we tried varying it in our study. 

We also made two experiments when the replacement of the stop words is allowed and not. We find that there is no big difference in the number of broken texts in either case. Furthermore, we experimented with various values of $k$ and the minimal USE score to find out how they affected the number of the attacked texts and the robustness of the XLM-RoBERTa model trained on them. Since the original TextFooler implementation in the TextAttack framework  \cite{textattack} does not contain embeddings for Russian, we used FastText embeddings for both English and Russian to make the experiments with both languages comparable. 

\autoref{tabZeroThreshold} shows that the number of the successfully attacked texts is practically independent from the USE threshold when it varies from the default 0.84 to 0, so this filter is not particularly useful for genre attacks. At the same time, the proportion of the broken texts increases when more variants for attack are considered (the value of $k$, the number of nearest neighbours to consider).

Besides, TextFooler turned out to be more efficient in breaking the Russian texts, about 15\% difference in the proportion of broken texts.  However, we should note that TextFooler tries to attack only texts which the model classifies correctly. As the XLM-RoBERTa classifier performed better on the Russian texts, we make more attacks on Russian texts in general.

Our experiments with applying TextFooler to genre classification produced convincing replacements which preserved the meaning for both English and Russian.  \autoref{tabUntargetedAttack} shows an example of a text, successfully broken by this mechanism in our task. A replacement of just one word in a reasonably long text is able to change the prediction of the classifier.  The words being replaced are typically not crucial for judging the genre.  Table~\ref{tabTextFoolerWordPairs} lists the most common word replacement pairs with untargeted attacks for each genre for English.  


We also found that preserving the grammar is trickier: \autoref{tabUngrammarUntargetedAttack} shows an example of alteration that makes a text ungrammatical. A word-level replacement mechanism is not enough to keep the sentence grammatical, as replacing $should\to need$ in this example requires syntactic alterations, and filtering by the Universal Sentence Encoder scores is not enough to guarantee syntactic coherence. Table~\ref{tabTextFoolerWordPairs} also shows that many replacements do not keep the grammatical number, for example, replacing $have\to has$, $opportunity\to opportunities$, which is likely to lead to ungrammatical sentences.  Also the replacements are not motivated, as often the replacements for the same genre can go in both directions, for example, $have\to be$ and $be\to have$, or they have nothing to do with the genres, for example, $Google\to Yahoo$.  We can assume that the lack of motivated replacements is coming from instability of parameters in the transformer model when small changes in the input texts lead to considerable changes in the output predictions.

Table~\ref{tabAttackAccuracy} lists the results for untargeted attack for both English and Russian FTD corpora in terms of the number of words needed for a successful change of genre predictions for a text.  The most difficult genres for attack in English are Legal, Fiction and Personal blogs.  This is likely because their training sets are less affected by topical biases.  In contrast, Information and Review texts are easier to attack with fewer substitutions, while they are more affected by topical biases, such as politics, see also the keywords in \autoref{tabKeywords} and the most salient replacement pairs in  \autoref{tabTextFoolerWordPairs}. However, the difference from tf-idf replacements is that TextFooler tends to amend more frequent English verbs, while the words chosen by the tf-df mechanism are more genre-specific. That confirms the observation that replacements in successful TextFooler attacks on genre classification are not motivated by genre if the original training set is biased.

\subsection{Targeted attacks with TextFooler}
\label{secTargetedTextFooler}

For targeted attacks we use the same mechanism with TextFooler, but we choose the replacement candidate that maximises the probability of the true class. 

\begin{table*}[t]
  \centering
\begin{tabular}{c|c|c|c} 
\hline
\textbf{Language} & \textbf{k=15} & \textbf{k=30} & \textbf{k=50} \\
\hline
EN & 233 (34.2\%) & 248 (36.4\%) & 254 (37.2\%) \\
\hline
RU & 317 (57.3\%) & 326 (59.0\%) & 328 (59.3\%) \\ 
\hline
\end{tabular}
\caption{The number of the texts broken by the targeted attack, USE threshold = 0.84}
\label{tabTargetedAttacks}
\end{table*}

\autoref{tabTargetedAttacks} lists for how many texts the classifier predictions can be improved by the attack mechanism. Targeted attacks are considerably harder that the untargeted ones.

\begin{table*}[!t]
  \centering
 \begin{tabular}{l|cc|cc|cc} 
\textbf{Genre} & 
\multicolumn{2}{c}{F1} & \multicolumn{2}{c}{Prec} &
\multicolumn{2}{c}{Rec}\\
& Base & Robust & Base & Robust &  Base & Robust \\
 \hline
Argument 	 & \textbf{0.585} & 0.550 & 0.514 & \textbf{0.612} &  \textbf{0.678} & 0.499 \\ 
Fiction 	 & \textbf{0.685} & 0.677 & \textbf{0.902} & 0.697 & 0.553 & \textbf{0.658} \\
Instruction  & 0.651 & \textbf{0.738} & \textbf{0.891} & 0.762 & \textbf{0.813} & 0.716 \\
News    	 & \textbf{0.940} & 0.937 & 0.917 & \textbf{0.943} & \textbf{0.965} & 0.931 \\
Legal   	 & 0.585 & \textbf{0.615} & 0.444 & \textbf{0.480} & \textbf{0.857} & \textbf{0.857} \\
Personal 	 & \textbf{0.742} & 0.723 & \textbf{0.747} & 0.657 & 0.737 & \textbf{0.805} \\
Promotion 	 & 0.333 & \textbf{0.408} & 0.316 & \textbf{0.369} & 0.353 & \textbf{0.456} \\
Academic 	 & 0.273 & \textbf{0.489} & 0.250 & \textbf{0.440} & 0.300 & \textbf{0.550} \\
Information  & \textbf{0.586} & 0.578 & \textbf{0.690} & 0.596 & 0.509 & \textbf{0.561} \\
Review  	 & \textbf{0.571} & 0.535 & 0.550 & \textbf{0.559} 	& \textbf{0.595} & 0.514 \\
 \end{tabular}
\caption{Comparison of the Base and the Robust XLM-RoBERTa results for English}
\label{tabComparisonEN}
\end{table*}

\begin{table*}[!t]
  \centering
 \begin{tabular}{l|cc|cc|cc} 
\textbf{Genre} & 
\multicolumn{2}{c}{F1} & \multicolumn{2}{c}{Prec} &
\multicolumn{2}{c}{Rec}\\
& Base & Robust & Base & Robust &  Base & Robust \\
 \hline
Argument     & 0.584 & \textbf{0.728} & 0.487 & \textbf{0.734} & \textbf{0.723} & \textbf{0.723} \\ 
Fiction 	 & 0.913 & \textbf{0.928} & \textbf{0.977} & 0.907	& 0.858 & \textbf{0.950} \\
Instruction  & 0.535 & \textbf{0.617} & \textbf{0.708} & 0.635 & 0.430 & \textbf{0.600} \\
News         & 0.816 & \textbf{0.848} & 0.710 & \textbf{0.767} & \textbf{0.960} & 0.948 \\
Legal   	 & \textbf{0.860} & 0.652 & 0.990 & \textbf{0.995} & \textbf{0.760} & 0.485 \\
Personal 	 & 0.682 & \textbf{0.707} & \textbf{0.719} & 0.685 & 0.648 & \textbf{0.730} \\
Promotion 	 & 0.819 & \textbf{0.881} & \textbf{0.914} & 0.912 & 0.742 & \textbf{0.853} \\
Academic 	 & \textbf{0.892} & 0.886 & \textbf{0.823} & 0.816 & \textbf{0.975} & 0.968 \\
Information  & \textbf{0.942} & 0.845 & \textbf{0.923} & 0.753 & \textbf{0.963} & \textbf{0.963} \\
Review  	 & \textbf{0.812} & 0.774 & \textbf{0.892} & 0.857 & \textbf{0.745} & 0.705 \\
 \end{tabular}
\caption{Comparison of the Base and the Robust XLM-RoBERTa results for Russian}
\label{tabComparisonRU}
\end{table*}

\begin{table*}
  \centering
 \begin{tabular}{c|c|c|c|c}
    \hline
    \textbf{Corpus} & \textbf{no attacked} & \textbf{k=15} & \textbf{k=30} & \textbf{k=50} \\
    \hline
    En, Natural & 0.747 $\pm$ 0.026 & 0.796 $\pm$ 0.011 & 0.771 $\pm$ 0.01 & 0.776 $\pm$ 0.029 \\
    \hline
    Ru, LiveJournal &  0.76 $\pm$ 0.003 & 0.756 $\pm$ 0.008 & 0.755 $\pm$ 0.009 & 0.756 $\pm$ 0.005 \\ 
    \hline
 \end{tabular}
 \caption{Accuracy of the XLM-RoBERTa classifier trained on the attacked texts}
\label{tabRobustAccuracy}
\end{table*}

\begin{table*}[]
\centering
 \begin{tabular}{l|cc|cc|cc} 
\textbf{Genre} & 
\multicolumn{2}{c}{F1} & \multicolumn{2}{c}{Prec} &
\multicolumn{2}{c}{Rec}\\
& Base & Robust & Base & Robust &  Base & Robust \\
 \hline
Argument    & 0.566 & \textbf{0.732} & 0.534 & \textbf{0.724} & 0.603       & \textbf{0.740}           \\
Fiction    & 0.914 & \textbf{0.929} & \textbf{0.951} & 0.913 & 0.88         & \textbf{0.945}          \\
Instruction    & 0.448 & \textbf{0.621} & 0.613 & \textbf{0.636} & 0.353       & \textbf{0.608}         \\
News    & 0.689 & \textbf{0.856} & 0.529 & \textbf{0.784} & \textbf{0.988}       & 0.943         \\
Legal   & \textbf{0.798} & 0.652 & 0.985 & \textbf{0.995} & \textbf{0.670}         & 0.485  \\
Personal   & 0.658  & \textbf{0.702}  & 0.580 & \textbf{0.681} & \textbf{0.760}         & 0.725          \\
Promotion   & 0.502 & \textbf{0.885} & 0.802 & \textbf{0.915} & 0.365        & \textbf{0.858}         \\
Academic   & \textbf{0.910}  & 0.888 & \textbf{0.883} & 0.820 & 0.940         & \textbf{0.968}         \\
Information  & \textbf{0.944} & 0.847 & \textbf{0.917} & 0.753 & \textbf{0.973}       & 0.968         \\
Review  & 0.752 & \textbf{0.777} & \textbf{0.933} & 0.865 & 0.630         & \textbf{0.705}          
\end{tabular}
\caption{Comparison of the Base and the Robust XLM-RoBERTa results for the English natural annotation corpus}
\label{tabENHomogenousComparison}
\end{table*}

\subsection{Adding attacked texts to train new genre classifiers}
\label{secReliableModel}

In the next step we add broken texts with correct labels to train a new model and we test it on the validation portion of the original training corpus and also on test corpora.  \autoref{tabRobustAccuracy} lists the robust classifier performance on the test corpora. It shows that the XLM-RoBERTa classifier trained on the attacked texts attains higher accuracy than the baseline classifier.  \autoref{tabENHomogenousComparison} shows, that for most genres the robust classifier achieves higher f1-score. The same is true for precision and recall. 



Training XLM-RoBERTa on concatenation of the original and broken texts does not improve the classifier performance on the LiveJournal corpus but significantly increases the accuracy on the English genre corpus with natural annotation. Besides, the best result is attained when hyper-parameter value $k=15$ is used. It shows that the quality of attack is more important than the number of the successfully attacked texts for boosting the classifier performance. In the \autoref{tabComparisonEN} we can see that the robust classifier performs better for most genres. In the  \autoref{tabComparisonRU} the improvement in terms of the F1 score is limited, since for many genres improving recall implies deterioration of precision. 


\section{Related Work}
\label{secRelatedWork}

Genre classification is not a new task, since non-topical classification is needed for many applications. There have been experiments with various  architectures from linear discriminant analysis \cite{karlgren94} to SVM \cite{dewdney01} to recurrent neural networks \cite{kunilovskaya19}.  Early work on robust genre classification across different training and testing corpora \cite{sharoff10lrec} reveals the problem of topical biases in the genre corpora available at the moment. In this paper we try to solve the problem indirectly by improving their robustness. \cite{petrenz10} investigate a very important idea concerning estimation of the reliability of genre classifiers via its validation on a corpus with different topical distributionss but with the same genre labels.  Our study continues this line of research when we use the datasets from natural annotation and LiveJournal to estimate the model accuracy on an out-of-domain testing corpus.

Our experiments with using adversarial attacks for genre classification are novel. The most efficient adversarial attack techniques for classifiers \cite{textfooler,bertattack} are based on usage of word-level embeddings and finding for each word a fixed number of the most similar words as candidates for replacing with. Our genre attacks are based on the TextFooler \cite{textfooler} with a modification that we allow replacing of the stop-words and vary the USE threshold. TextFooler \cite{textfooler} was chosen as the basis for genre attacks in this study due to its efficiency and flexibility as it can be applied to various neural models. We also experimented with BertAttack, that differs from the TestFooler algorithm in its use of BERT token embeddings instead of pre-trained word-level embeddings.  In our initial experiments we found it to be much slower than TextFooler and also somewhat less efficient for the genre attack task, for example, the percentage of the texts successfully broken by BertAttack is lower than 15\% for English. Therefore, we only report the results with TextFooler here.  A recent experiment on adversarial attacks on personal style, as another non-topical classification task, \cite{emmery21profiling} is the closest to our study.  They attacked author profile predictions using similar methods.  However, they have not investigated the question of attacks on genre predictions.

\section{Conclusion}
\label{secConclusions}

In this paper we show that the XLM-RoBERTa genre classifier is resistant to simple attack methods, such as replacement of tf-idf keywords, this is unlike traditional feature-based methods which are very sensitive to the keywords.  At the same time, even the XLM-RoBERTa classifier can be deceived by word-based adversarial attacks using mechanisms like TextFooler. In the case of the baseline classifier, more than 35\% of English texts in the training corpus can be successfully broken, raising to more than 50\% for Russian.  The number of successfully attacked texts can be considered as an important metric for estimating the robustness of the classifiers -- the lower the number of broken texts, the more difficult it is to break the classifier, which implies higher robustness.  Also we find some important patterns in the attack results, in particular, the threshold for USE almost does not affect the number of the attacked texts; attacks are more efficient for the Russian language; the higher the number of replacing candidates, the less the difference in reliability of the robust classifier vs the original one. 

Our experiments demonstrate the  effectiveness  of TextFooler at improving robustness of genre classifiers via adversarial attacks. For example, adding broken texts (with their original labels) improves the overall accuracy, while texts in the new collection cannot be broken by the same set of adversarial attacks, thus implying a more robust classifier.  We also tried targeted attacks, but fewer text can be broken and the classifiers trained on the targeted attacked texts performed worse than those coming from the untargeted attack.


\bibliographystyle{apalike}
\bibliography{bibexport}

\end{document}